\documentclass[conference]{IEEEtran}
\IEEEoverridecommandlockouts

\usepackage{cite}
\usepackage{amsmath,amssymb,amsfonts}
\usepackage{algorithmic}
\usepackage{graphicx}
\usepackage{textcomp}
\usepackage{xcolor}
\usepackage{booktabs}
\usepackage{multirow}
\usepackage{multicol}
\usepackage{hyperref}
\usepackage{url}
\def\BibTeX{{\rm B\kern-.05em{\sc i\kern-.025em b}\kern-.08em
    T\kern-.1667em\lower.7ex\hbox{E}\kern-.125emX}}

\begin{document}

\title{Advancing Scientific Text Classification: Fine-Tuned Models with Dataset Expansion and Hard-Voting}

\author{\IEEEauthorblockN{Zhyar Rzgar K Rostam}
\IEEEauthorblockA{\textit{Doctoral School of Applied Informatics and Applied Mathematics} \\
\textit{Obuda University}\\
Budapest, Hungary \\
zhyar.rostam@stud.uni-obuda.hu}
\and
\IEEEauthorblockN{Gábor Kertész}
\IEEEauthorblockA{\textit{John von Neumann Faculty of Informatics} \\
\textit{Obuda University}\\
Budapest, Hungary \\
kertesz.gabor@nik.uni-obuda.hu}

}

\maketitle

\begin{abstract}
Efficient text classification is essential for handling the increasing volume of academic publications. This study explores the use of pre-trained language models (PLMs), including BERT, SciBERT, BioBERT, and BlueBERT, fine-tuned on the Web of Science (WoS-46985) dataset for scientific text classification. To enhance performance, we augment the dataset by executing seven targeted queries in the WoS database, retrieving 1,000 articles per category aligned with WoS-46985’s main classes. PLMs predict labels for this unlabeled data, and a hard-voting strategy combines predictions for improved accuracy and confidence. Fine-tuning on the expanded dataset with dynamic learning rates and early stopping significantly boosts classification accuracy, especially in specialized domains. Domain-specific models like SciBERT and BioBERT consistently outperform general-purpose models such as BERT. These findings underscore the efficacy of dataset augmentation, inference-driven label prediction, hard-voting, and fine-tuning techniques in creating robust and scalable solutions for automated academic text classification.

\end{abstract}


\section{Introduction}

The rapid growth of academic literature has led to a significant increase in the volume of research articles, conference papers, and books \cite{fields2024survey, borrajo2011improving, luo2023exploring, app13063594, ahanger2022novel, naseem2022benchmarking, jiao2023brief}. Effectively managing and organizing this large volume of data is a challenging task. Manual classification is time-consuming and prone to errors due to fatigue and subjectivity \cite{app13063594}. To address this, Natural Language Processing (NLP) techniques have become essential for automating text classification and accurately categorizing research content into specialized domains. \cite{minaee2021deep}. Deep learning (DL), particularly Large Language Models (LLMs) based on transformer architectures, has achieved remarkable success in tasks like text classification \cite{minaee2021deep, naseem2022benchmarking, fields2024survey, peng2021survey, jiao2023brief, sun2023text}, sentiment analysis \cite{alimova2021cross, araci2019finbert, laki2023sentiment}, topic modeling \cite{kherwa2019topic, lezama2023integrating}, information retrieval \cite{zhao2023survey, peng2021survey}, and natural language inference.

However, general-purpose LLMs often struggle with domain-specific tasks \cite{sun2023text, vaswani2017attention}, such as scientific research classification, due to challenges like specialized vocabulary, unique grammatical patterns, and imbalanced data distributions. \cite{beltagy2019scibert, lee2020biobert, rostam2024fine, devlin2018bert, liu2021finbert, dunn2022structured, gupta2022matscibert}. These factors can reduce the efficacy of general-purpose LLMs for tasks like scientific text classification, highlighting the need for domain-specific model adaptations and data augmentation to enhance performance in domain-specific contexts \cite{devlin2018bert} \cite{beltagy2019scibert, lee2020biobert, peng2019transfer, liu2021finbert, kim2023medibiodeberta}.

In this study, we enhance model generalization and achieve better accuracy through dataset expansion and label prediction. By applying seven distinct queries to the Web of Science (WoS) database, we retrieved 1,000 articles across seven academic domains (details presented in Table \ref{tab:original_constructed_ds}). Using four fine-tuned pre-trained language models ($BERT_{base}$ \cite{devlin2018bert}, $SciBERT_{scivocab}$ \cite{beltagy2019scibert}, $BioBERT_{base}$ \cite{lee2020biobert}, and $BlueBERT_{large}$ \cite{peng2019transfer}), we generated labels for the retrieved articles and employed a hard-voting strategy to consolidate predictions with high confidence. \\
Our results demonstrate that this dataset expansion approach improves classification accuracy and generalization, outperforming existing methods. This research highlights the effectiveness of combining domain-specific LLM fine-tuning with strategic data augmentation, providing a scalable solution for automated scholarly content classification \footnote{All datasets, models fine-tuning results can be accessed at: \url{https://github.com/ZhyarUoS/Advancing-Scientific-Text-Classification.git}}. The contributions of this study are:
\begin{itemize}
    \item Expand the dataset by adding additional articles to enhance model generalization.
    \item Apply a hard-voting strategy to combine predictions and increase classification accuracy by ensuring high-confidence label assignment.
    \item Outperform baseline models and previous studies in domain-specific text classification tasks.
    \item Optimize training with dynamic learning rates and early stopping to reduce computational cost.
    \item Fine-tune PLMs on the expanded domain-specific datasets and provide a detailed comparison with models stated in the literature.
\end{itemize}

\section{Related Works}

\subsection{Data Augmentation with LLMs for Text Classification}
Danilov et al. \cite{danilov2021classification} investigate automated text classification methods for classifying scientific context, specifically using a dataset of 630 PubMed abstracts in binary classification tasks. Researchers fine-tuned BERT model and experimented with different setups to improve classification accuracy. They also explored ensemble models, merging predictions from PubMedBERT, logistic regression, random forest, and support vector machines. Their results demonstrate that both PubMedBERT and ensemble models achieved remarkable success in classifying short scientific texts. 

Jaradat et al. \cite{jaradat2024ensemble} present a novel crash severity classification approach that combines machine learning and deep learning techniques. The study propose an ensemble voting classifier that integrates PLMs (BERT and RoBERTa), fine-tuned with Bi-LSTM, alongside traditional machine learning models (XGBoost, random forest, and naive Bayes) and word embedding methods (TF-IDF, Word2Vec, and BERT). Through extensive experiments on a real-world crash dataset, the study demonstrates the effectiveness of this approach in improving classification accuracy, particularly for the minority class. The findings show the importance of data augmentation and ensemble learning in addressing class imbalance and enhancing model performance.

\subsection{Deep Learning Approaches and Data Augmentation for Scientific Text Classification}
Wei and Zou \cite{wei2019eda} introduce a simple effective data augmentation method, easy data augmentation (EDA), aimed to enhance text classification model performance by expanding training dataset using four strategies: synonym replacement, random insertion, random swap, and random selection. These techniques lead to notable performance enhancements, especially on smaller datasets. The effectiveness of EDA is consistently observed among various benchmark datasets and model types (including RNNs and CNNs). The study reveals that EDA is particularly applicable when working with limited training data, underscoring its value for data-scarce scenarios.

HDLTex \cite{kowsari2017HDLTex} is a hierarchical deep learning model designed to classify large and complex document collections. This model effectively categorizes documents into multiple levels of categories by employing a combination of deep learning architectures, including Deep Neural Networks (DNNs) \cite{schroder2020survey}, Convolutional Neural Networks (CNNs) \cite{lai2015recurrent}, and Recurrent Neural Networks (RNNs) \cite{lee2016sequential, lai2015recurrent}. This approach enables the model to handle detailed patterns in text data across multiple category levels.

\section{Dataset}
\label{dataset}

\subsection{Original Dataset}
\label{original_dataset}

We utilize the Web of Science (WoS) dataset, originally collected by Kowsari et al. \cite{kowsari2017HDLTex}, as the baseline for our study. This dataset comprises three main subsets, each varying in size and categorization: WoS-5736, WoS-11967, and WoS-46985 (presented in Tables \ref{tab:original_constructed_ds}, \ref{tab:11967_abstracts} and \ref{tab:wos_5736}, respectively). The smallest subset, WoS-5736, contains 5,736 documents categorized into 11 subcategories within three primary scientific domains. WoS-11967 is larger, with 11,967 documents across 35 subcategories under seven scientific domains, while WoS-46985, the largest, includes 46,985 documents and spans 134 subcategories within the same seven domains.

\subsection{Constructed Dataset}
\label{original_dataset}
To improve classification accuracy and generalization, we expanded the original WoS datasets by retrieving additional documents from the WoS database using seven targeted queries across key categories defined by Kowsari et al. \cite{kowsari2017HDLTex}. From each query, we extracted the top 1,000 articles per domain, with duplicates removed. Multi-model inference and hard-voting were used to assign labels based on model agreement. The resulting expanded datasets WoS-8716, WoS-18932, and WoS-53949, (presented in Tables \ref{tab:original_constructed_ds}, \ref{tab:11967_abstracts} and \ref{tab:wos_5736}, respectively), contain 8,715, 18,932, and 53,949 unique documents, respectively. This expanded dataset provides a more comprehensive representation of each category, enabling the development of more accurate and reliable classification models.

\begin{table}[tbp]
\caption{Number of studies per category in both the original (WoS-46985) and expanded (WoS-53949) datasets, with the expanded dataset totals reflecting hard-voting results}
    \centering
    \begin{tabular}{ l  c  c}
        \hline
        \textbf{Domain} & \textbf{WoS-46985 \cite{kowsari2017HDLTex}} & \textbf{WoS-53949} \\ \hline
        Computer Science & 6514 & 7773 \\ \hline
        Civil Engineering & 4237 & 5000 \\ \hline
        Electrical Engineering & 5483 & 6303 \\ \hline
        Mechanical Engineering & 3297 & 4472 \\ \hline
        Medical Sciences & 14625 & 15641\\ \hline
        Psychology & 7142 & 8089\\ \hline
        Biochemistry & 5687 & 6671\\ \hline
        \textbf{Total} & \textbf{46985} &  \textbf{53949} \\ \hline
    \end{tabular}
    \label{tab:original_constructed_ds}
\end{table}

\begin{table}[tbp]
 \caption{Number of studies per category in both the original (WoS-11967) and expanded (WoS-18932) datasets, with the expanded dataset totals reflecting hard-voting results}
    \centering
    \begin{tabular}{ l  c c}
        \hline
        \textbf{Domain} & \textbf{WoS-11967 \cite{kowsari2017HDLTex}} & \textbf{WoS-18932} \\ \hline
        Computer Science & 1499 & 2758\\ \hline
        Civil Engineering & 2107 & 2870\\ \hline
        Electrical Engineering & 1132 & 1952\\ \hline
        Mechanical Engineering & 1925 & 3100\\ \hline
        Medical Sciences & 1617 & 2634\\ \hline
        Psychology & 1959 & 2906\\ \hline
        Biochemistry & 1728 & 2712\\ \hline
        \textbf{Total} & \textbf{11967} & \textbf{18932} \\ \hline
    \end{tabular}
    \label{tab:11967_abstracts}
\end{table}

\begin{table}[tbp]
    \caption{Number of studies per category in both the original (WoS-5736) and expanded (WoS-8716) datasets, with the expanded dataset totals reflecting hard-voting results}
    \centering
    \begin{tabular}{ l  c c}
        \hline
        \textbf{Domain} &  \textbf{WoS-5736 \cite{kowsari2017HDLTex}} & \textbf{WoS-8716}\\ \hline
        Electrical Engineering & 1292 & 2245\\ \hline
        Psychology & 1597 & 2632\\ \hline
        Biochemistry & 2847 & 3839\\ \hline
        \textbf{Total} & \textbf{5736} &  \textbf{8716}\\ \hline
    \end{tabular}
    \label{tab:wos_5736}
\end{table}

\section{Methods}

\subsection{Models Inference}
\label{inferance}
We utilized the fine-tuned models (BERT, SciBERT, BioBERT, and BlueBERT) obtained from the baseline study by Rostam and Kertész \cite{rostam2024fine} to perform document classification and category prediction for retrieved unseen datasets. For each document, titles, abstracts, and keywords are cleaned and combined as input, then passed to the model for inference. The model generates probabilistic predictions across categories, providing a ranked list of the top three categories for each document. Table \ref{tab:models_inferance} presents a comparison for model predictions against query classes.

\subsection{Hard-voting and Models Agreement}
\label{hard-voting}
\begin{figure}[tbp]
\centerline{\includegraphics[width=.5\textwidth]{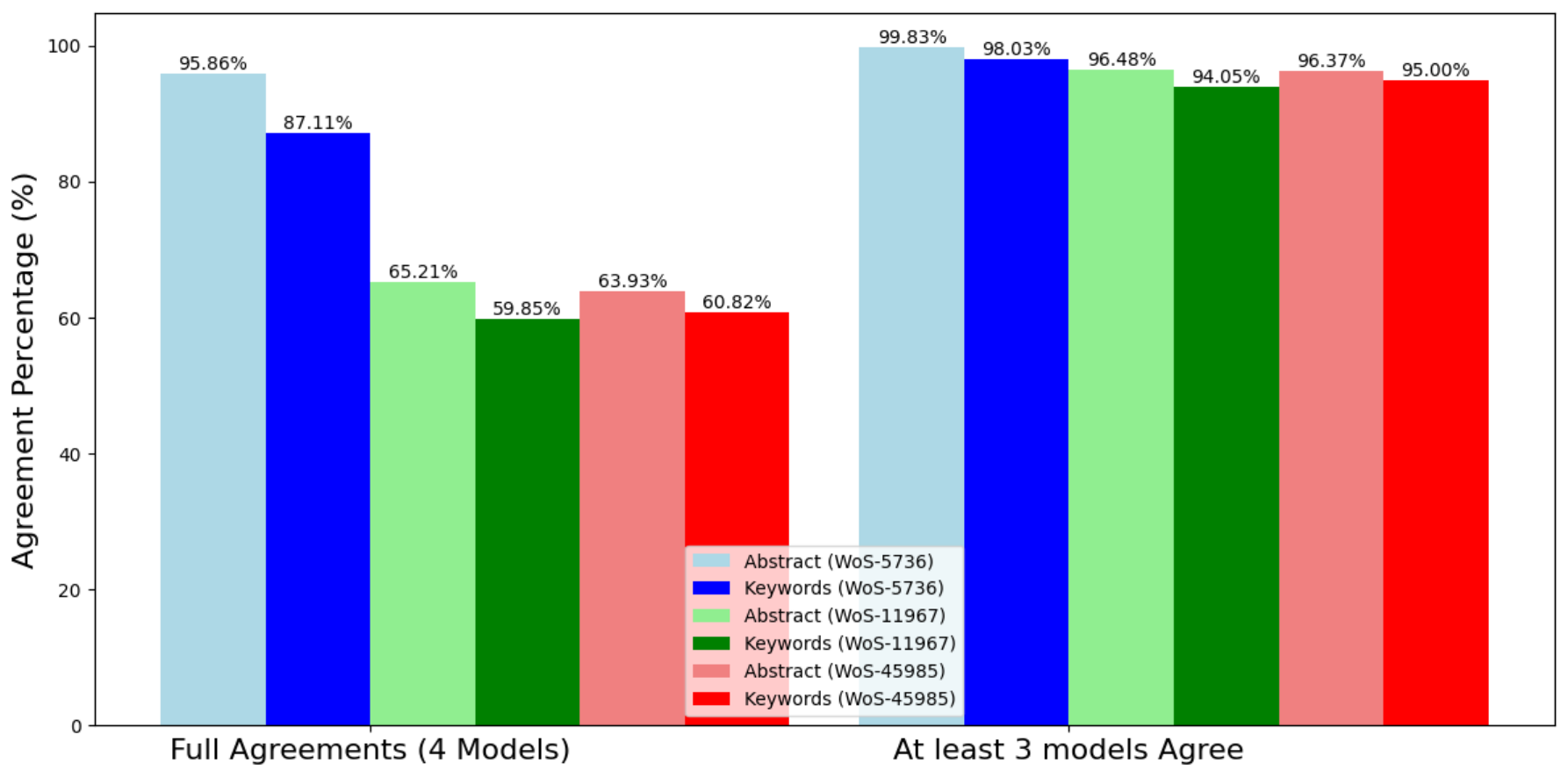}}
\caption{PLMs agreement results across both scenarios (abstract and keywords)}
\label{fig:full_models_agreements}
\end{figure}

We ranked the prediction probabilities generated by the PLMs for individual studies and select the highest probability prediction from each model. Hard-voting mechanism is utilized to achieve agreement among the models for labeling individual studies. This procedure produced a newly labeled dataset, which we combined with the original WoS dataset versions as a result three different datasets (WoS-53949, WoS-18932, and WoS-8716) as presented in Table \ref{tab:original_constructed_ds}, \ref{tab:11967_abstracts} and \ref{tab:wos_5736}) constructed. The results of all models agreement in both scenarios (abstract, and keywords) are presented in Figure: \ref{fig:full_models_agreements}.

\subsection{Experimental Design}
\label{expermintal_design}
This study focused on fine-tuning PLMs for scientific text classification, using both general-purpose (BERT) and domain-specific (SciBERT, BioBERT, and BlueBERT) models. The aim was to evaluate their performance in handling scientific text and identifying the most effective approaches for the task.

\subsection{Data Preperation}
\label{data_preperation}
To prepare the textual data, comprising abstracts and keywords, for fine-tuning, we performed tokenization and encoding processes. These steps were essential to convert raw text into a numerical representation suitable with the PLMs. To ensure consistency and reduce noise in the dataset, we applied text cleaning by converting all the texts to lowercase and removing non-alphabetic characters. Following this, we utilized model-specific tokenizer associated with the PLMs (BERT, SciBERT, BioBERT, BlueBERT) to tokenize the text into smaller units aligned with the tokenizer's vocabulary. Then, the tokenized texts were converted into numerical representations. Specifically, input IDs were generated to represent tokens as unique integers, while attention masks were created to distinguish relevant tokens from padding.\\
To ensure generalization and mitigate potential biases, the dataset was shuffled prior to splitting, The dataset was divided into training (80\%), testing (20\%), and validation (50\% of the testing set, which is equal to 10\% of the dataset) as presented in Table \ref{tab:dataset_splits}.
\begin{table}[tbp]
    \centering
    \caption{Dataset Splits for WoS Datasets}
    \label{tab:dataset_splits}
    \begin{tabular}{lrrr}
        \toprule
        \textbf{Dataset} & \textbf{Train} & \textbf{Test} & \textbf{Validation} \\
        \midrule
        WoS-8716  & 6972  & 1744  & 872  \\
        WoS-18932 & 15145  & 3787  & 1894  \\
        WoS-53949 & 43159 & 10790  & 5395 \\
        \bottomrule
    \end{tabular}
\end{table}

\subsection{Fine-tuning Procedure}
\label{fine_tuning}
We followed a standardized fine-tuning process to ensure a fair comparison. The models were fine-tuned over a maximum of 20 epochs using AdamW optimizer, configured with an epsilon  \(1 \times 10^{-8}\) to ensure numerical stability. A dynamic learning rate search strategy was utilized to identify the optimal learning rate from a set of values (\(2 \times 10^{-5}\), \(5 \times 10^{-6}\), \(1 \times 10^{-6}\), \(2 \times 10^{-6}\)). Additionally, a linear warmup scheduler, followed by gradual decay, was utilized to stabilize learning and improve convergence (the details presented in Table \ref{tab:training_parameters}). To avoid overfitting and improve generalization, an early stopping mechanism was implemented. This strategy monitored the validation F1 score and terminated the process if no improvement was observed. Finally, the model corresponding to the highest validation F1 score was saved and subsequently used for final testing.

\begin{table}[tbp]
    \centering
    \caption{Training Configuration Parameters}
    \label{tab:training_parameters}
    \begin{tabular}{lr}
        \toprule
        \textbf{Parameter} & \textbf{Value} \\
        \midrule
        Optimizer          & AdamW \\
        Learning Rates      & \{\(2 \times 10^{-5}\), \(5 \times 10^{-6}\), \(1 \times 10^{-6}\), \(2 \times 10^{-6}\)\} \\
        Epsilon            & \(1 \times 10^{-8}\) \\
        Scheduler          & Linear with warmup \\
        Warmup Steps       & \(1 \times 10^{-4}\) \\
        Epochs & \( \text{20} \) \\
        \bottomrule
    \end{tabular}
\end{table}

\section{Results}
In this section, we present the model's performance and efficiency for inference on the unlabeled data (presented in Table \ref{tab:models_inferance}) and fine-tuned results on the constructed dataset after giving labels based on the model's agreements. Finally, we provide a comparison against the best-achieved results reported in the literature among our experimented results (see Table \ref{tab:models_accuracies}). All models' performance evaluations are presented in Table \ref{tab:model_perfomance_metrics}.

\subsection{Inference on Unlabeled Data: Query Classes vs. Predictions (WoS-5736)}
As presented in Table \ref{tab:wos_5736}, WoS-5736 comprises three classes: psychology, biochemistry, and ECE. Therefore, we conducted inference exclusively for these categories. All models performed well when fed with abstracts, with SciBERT leading in biochemistry (97.70\%) and BERT excelling in ECE (96.39\%). When keywords were used, BERT achieved exceptional performance in biochemistry, while SciBERT dominated in psychology and ECE. Despite its strong results in psychology and ECE, BlueBERT struggled in biochemistry with keywords (73.60\%). These results highlight the adaptability of SciBERT and BERT across domains and input types.

\subsection{Inference on Unlabeled Data: Query Classes vs. Predictions (WoS-11967)}
PLMs fine-tuned on abstracts showed varied performance across domains. BERT excelled in MAE and biochemistry (98.0\%, 92.20\%) but struggled in CS and medical sciences. SciBERT, designed for scientific texts, performed well in MAE and biochemistry (96.60\%, 94.0\%) but lagged in medical sciences and CS. BioBERT and BlueBERT, tuned for biomedical contexts, delivered strong results in MAE and biochemistry but faced challenges in CS. For models fine-tuned with keywords, SciBERT excelled in CS and biochemistry (92.0\%, 96.60\%), while BERT performed well in biochemistry and MAE (94.20\%, 97.10\%) but struggled in CS and medical sciences. BioBERT achieved consistent performance across domains, excelling in CS and biochemistry (95.40\%, 92.20\%), while BlueBERT showed strong results in biochemistry and MAE. These findings underscore SciBERT and BioBERT's strengths in scientific and biomedical contexts (see Table \ref{tab:models_inferance}).

\subsection{Inference on Unlabeled Data: Query Classes vs. Predictions (WoS-46985)}
All four models demonstrated exceptional accuracy in CS, and strong performance in MAE, ranging from 75.0\% to 84.00\%. However, their accuracy dropped noticeably in more specialized fields such as medical sciences and biochemistry, where most models struggled to exceed 60\%, highlighting challenges in domain-specific generalization. When using models fine-tuned with keywords, an improved balance in performance emerged across domains. SciBERT excelled in psychology and MAE, reflecting its scientific pre-training advantages, while maintaining strong accuracy in biochemistry. BERT outperformed other models in CS and medical sciences domains. BioBERT and BlueBERT performed well in CS and psychology but struggled with civil engineering and biochemistry domains. These results highlight the strengths of keyword-based fine-tuning in enhancing domain-specific predictions, particularly for BERT and SciBERT (see Table \ref{tab:models_inferance}).

\begin{table}[tbp]
\caption{Inference Results for Unlabeled Data: Comparison of Model Predictions with Query Classes}
\label{tab:models_inferance}
\scalebox{0.7}
{
\begin{tabular}{l|l|l|l|l|l|l|l}
\toprule
\multirow{2}{*}{PLM}    & \multirow{2}{*}{Domain}    & \multicolumn{2}{c}{WoS-11967}       & \multicolumn{2}{c}{WoS-46985} & \multicolumn{2}{c}{WoS-5736} \\ \cline{3-8}
                       &      & Abstract & Keywords     & Abstract  & Keywords & Abstract  & Keywords \\
                       \hline
BERT  &   CS           & 91.10   &     34.60    & 100 & 99.90 & - & - \\   
        &   ECE           & 85.24 & 74.80   &     73.39          & 74.50 & 96.39 & 96.89 \\ 
        &   Psychology           & 58.36 & 64.26   &     60.86         & 65.47 & 99.80 & 98.30 \\ 
        &   MAE           & 98.0 &  97.10  &     84.0        & 75.80  & - & -\\ 
        &   Civil Eng.           & 68.72 & 50.95   &     72.87         & 73.77  & - & -\\ 
        &   Medical Sci.          & 57.20 &  37.60 &     61.20          & 76.70  & - & - \\ 
      &   Biochemistry        & 92.20 & 94.20 &     54.90        & 73.70  & 96.50 & 99.60\\ 
\midrule

SciBERT  &   CS           & 68.30 & 92.0 &     99.80          & 99.80 & - & -  \\   
        &   ECE           & 78.61 & 71.79 &     73.09          & 71.59  & 94.88 & 97.29\\ 
        &   Psychology           & 69.07 & 73.47    &     67.57          & 88.59 & 99.70 & 99.80 \\ 
        &   MAE           & 96.60 & 96.30    &     83.30          & 85.10  & - & -\\ 
        &   Civil Eng.           & 62.36 & 45.35    &     71.37          & 61.46 & - & - \\ 
        &   Medical           & 44.70 & 43.70  &     59.80          & 47.90 & - & -\\ 
      &   Biochemistry        & 94.00 & 96.60  &     47.90          & 74.90  & 97.70 & 94.60\\ 
\midrule

BioBERT  &   CS           & 49.30 & 95.40  &     99.60         & 99.60  & - & -\\   
        &   ECE           & 80.32 &  66.57&     70.98         & 68.70  & 95.88 & 92.97\\ 
        &   Psychology           & 57.16 & 65.17   &     70.07          & 83.48 & 99.10 & 99.50 \\ 
        &   MAE           & 96.10 & 96.60    &     75.00          & 68.0 & - & - \\ 
        &   Civil Eng.           & 58.46 & 40.04    &     76.48          & 56.26 & - & - \\ 
        &   Medical Sci.            & 58.10 & 78.10    &       54.0       & 65.50& - & -\\ 
      &   Biochemistry        & 91.50 & 92.20    &     53.40         & 56.90  & 95.70 & 92.90\\ 
\midrule

BlueBERT  &   CS           & 40.00 & 80.10    &     99.70          & 100.0  & - & -\\   
        &   ECE           & 82.03 & 75.60    &     78.92          & 69.98  & 95.28 & 96.08\\ 
        &   Psychology           & 66.87 & 76.58    &     62.26          & 84.68 & 99.60 & 99.20 \\ 
        &   MAE           & 96.10 & 95.20    &     83.70         & 74.50 & - & -\\ 
        &   Civil Eng.           & 67.57 & 63.56    &     73.37          & 59.66  & - & -\\ 
        &   Medical Sci.          & 63.10 & 57.80    &     53.50          & 51.30  & - & -\\ 
      &   Biochemistry        & 95.10 & 89.90    &     56.10         & 71.30  & 97.40 & 73.60\\ 
\midrule  
\end{tabular}
}
\end{table}

\subsection{Experimental Results: WoS-53949}
The experimental results on the WoS-53949 dataset demonstrated strong performance across all four PLMs. SciBERT emerged as the top-performing model, achieving micro F1 0.8924. The optimal LR \(1 \times 10^{-6}\) identified through early stopping, contributed to the model's peak validation micro F1 score of 0.8971. BioBERT achieved its peak validation micro F1 score of 0.8782 with LR of \(1 \times 10^{-6}\), while BERT reached its best validation micro F1 of 0.8788 at \(1 \times 10^{-6}\), demonstrating effectiveness with reduced training complexity and early stopping. BlueBERT achieved comparable results with micro F1 of 0.8828  (see Table \ref{tab:model_perfomance_metrics}).

\subsection{Experimental Results: WoS-18932}
In this experiment, BioBERT was able to outperform all other models in classification results by achieving the highest micro F1 0.9247. This result highlights BioBERT's superior capability in handling the WoS-18932 dataset. SciBERT, on the other hand, surpassed the other models in micro F1 (0.9244), demonstrating a balanced and comprehensive performance among all categories. This suggests SciBERT's effectiveness in balancing precision and recall, which is important for datasets with multiple classes. BERT and BlueBERT maintained strong performance but were slightly outperformed by the more specialized PLMs  (see Table \ref{tab:model_perfomance_metrics}). These results underline the importance of using domain-specific models for tasks like biomedical text classification, where molds such BioBERT, and SciBERT outperformed general purpose models like BERT.

\subsection{Experimental Results: WoS-5736}
For the WoS-5736 dataset, BERT achieved the highest in a final micro F1 score of 0.9782 and an accuracy of 98\%. SciBERT also performed well, reaching a micro F1 score of 0.9816. BioBERT demonstrated slightly more variability but still achieved a micro F1 score of 0.9816 and an accuracy of 98\%. BlueBERT had the lowest performance across the models, with a final micro F1 score of 0.9747 and an accuracy of 97\%  (see Table \ref{tab:model_perfomance_metrics}). Overall, due to the size and class distribution of the dataset, the models achieved better results compared to other datasets. This highlights the importance of domain-specific models for scientific classification tasks.

\begin{table}[tbp]
\centering
\caption{Models Performance Evaluation}
\label{tab:model_perfomance_metrics}
\scalebox{0.9}
{
\begin{tabular}{c|c|c|c|c}
\toprule

\textbf{Models} & \textbf{Micro F1} & \textbf{Micro Recall} & \textbf{Micro Precision} & \textbf{Accuracy}
\\
\midrule
\multicolumn{5}{c}{\textbf{WoS-53949}}
\\
\midrule

\textbf{BERT} &  0.8764 &  0.8587 &  0.8764 & 88\%\\
\midrule

 \textbf{SciBERT} &  \textbf{0.8924} &  \textbf{0.8924}&  \textbf{0.8924} & \textbf{89\%}\\
 
\midrule

\textbf{BioBERT} & 0.8792 &  0.8792 &  0.8792 & 88\%\\
\midrule

 \textbf{BlueBERT} &  0.8828 &  0.8828 & 0.8828 & 88\%\\
\bottomrule

\midrule
\multicolumn{5}{c}{\textbf{WoS-18932}}
\\
\midrule

 \textbf{BERT} & 0.9118 &  0.9118 &  0.9180 & 91\%\\
\midrule

\textbf{SCiBERT} &  0.9244 & 0.9244 &  0.9244 & \textbf{92\%}\\
\midrule

 \textbf{BioBERT} & \textbf{0.9247} &  \textbf{0.9247} &  \textbf{0.9247} & \textbf{92\%}\\
\midrule

\textbf{BlueBERT} &  0.9184 &  0.9184 &  0.9184 & \textbf{92\%}\\
 
\bottomrule


\midrule
\multicolumn{5}{c}{\textbf{WoS-8716}}
\\
\midrule

 \textbf{BERT} & 0.9782 &  0.9782 &  0.9782 & \textbf{98\%} \\
 
\midrule

\textbf{SciBERT} &  0.9816 &  0.9816 &  0.9816 & \textbf{98\%}\\
 
\midrule

 \textbf{BioBERT} &   \textbf{0.9827 }&   \textbf{0.9827} &   \textbf{0.9827} & \textbf{98\%} \\
 
\midrule

 \textbf{BlueBERT} &   0.9747 &   0.9747 &  0.9747 & 97\%\\
 
\bottomrule

\end{tabular}
}
\end{table}

\section{Discussion}
Our experiments demonstrate the model fine-tuned on domain-specific data, such as SciBERT and BioBERT, outperforms general purpose models like BERT among various scientific domains. The use of an expanded dataset significantly improved model performance, especially in specialized fields like biochemistry and medical sciences. The hard-voting strategy further enhanced classification accuracy by combining predictions from multiple models, ensuring more reliable results. This approach consistently outperformed baseline methods, including previous studies by Kowsari et al. \cite{kowsari2017HDLTex} and Rostam and Kertész \cite{rostam2024fine}.\\ 
Among the different models investigated, SciBERT performed best across most datasets, achieving the highest F1 scores in several categories. BERT showed strong performance in CS and medical sciences, while BioBERT excelled in biochemistry. Interestingly, BlueBERT, although specialized in biomedical text, fell behind other models in certain domains, highlighting the importance of selecting the right model for specific tasks. In comparison with previous works in the literature, our approach outperformed various deep learning models, including HDLTex, and demonstrated superior performance compared to general-purpose LLMs when fine-tuned with keywords and abstracts. The results underscore the benefits of leveraging domain-specific fine-tuning and dataset augmentation to enhance model generalization, providing a more robust and scalable solution for classifying academic content (see Table \ref{tab:models_accuracies}).

\begin{table}[tbp]
\caption{LLMs Accuracy (\%) Against Other Deep Learning Approaches}
\label{tab:models_accuracies}
\begin{tabular}{ll|c|c|c}
\toprule
\textbf{Models}    &  \textbf{Methods}      & \textbf{WoS-11967}               & \textbf{WoS-46985} & \textbf{WoS-5736} \\
                             
 \midrule                            
Baseline &   DNN           & 80.02             &     66.95          & 86.15  \\
                             &   CNN      &  83.29            & 70.46              & 88.68 \\
                             &   RNN            &  83.96            &   72.12            & 89.46  \\
                             &   NBC          &      68.8        &       46.2        & 78.14         \\
                             &   SVM            &  80.65            &   67.56            & 85.54  \\
                             &   SVM           &    83.16          &    70.22           & 88.24  \\
                             &   Stacking SVM          &   79.45           &     71.81          & 85.68       \\
\midrule                             
HDLTex \cite{kowsari2017HDLTex}                        &  HDLTex  &  86.07          &  76.58            &    90.93     \\
 \midrule                      
 \textbf{Models}    &  \textbf{Methods}      & \textbf{WoS-46985}               & \textbf{WoS-11967} & \textbf{WoS-5736} \\
  \midrule  
LLMs:  &   BERT           & 85.0             &     91.0          & 96.0  \\   
Abstracts \cite{rostam2024fine}&   SciBERT        & \textbf{87.0}             &     \textbf{92.0}          & 97.0  \\ 
                 &   BioBERT        & 86.0             &     91.0          & \textbf{98.0}  \\
                 &   BlueBERT       & 86.0             &     91.0          & 97.0  \\ 
\midrule 
LLMs:   &   BERT           & 79.0             &     84.0          & 93.0  \\   
Keywords \cite{rostam2024fine}   &   SciBERT        & \textbf{80.0}             &     \textbf{87.0}          & \textbf{94.0}  \\ 
                 &   BioBERT        & 79.0             &     86.0          & 93.0  \\
                 &   BlueBERT       & \textbf{80.0}             &     85.0          & 93.0  \\  
 \bottomrule

\midrule 
   &       & \textbf{WoS-53949}              & \textbf{WoS-18932} & \textbf{WoS-8716}\\ \cline{3-5} 
\textbf{LLMs(Ours):}   &   BERT           & 88.0             &     91.0          & \textbf{98.0}  \\   
\textbf{Abstracts and }  &   SciBERT        & \textbf{89.0}            &     \textbf{92.0}          & \textbf{98.0}  \\ 
\textbf{Keywords} &   BioBERT        & 88.0             &     \textbf{92.0}          & \textbf{98.0}  \\
             &   BlueBERT       & 88.0            &    \textbf{92.0}         & 97.0  \\ 
\midrule 

\end{tabular}
\end{table}

\section{Conclusion and Future Directions}
This study investigated the performance of both general-purpose (BERT) and domain-specific (SciBERT, BioBERT, and BlueBERT) PLMs on scientific texts. We expanded the WoS-46985 dataset by executing seven queries on the WoS database, each associated with the main categories in WoS-46985, and extracting the top 1,000 retrieved articles per category. We then performed inference using fine-tuned PLMs referenced in the literature to predict labels for the unlabeled data, employing a hard-voting strategy to ensure high-confidence label assignment. The most-voted labels were assigned, and the newly labeled data was integrated with the original WoS-46985 dataset, resulting in three expanded versions: WoS-8716, WoS-18932, and WoS-53949. Subsequently, PLMs were fine-tuned on these augmented datasets, which included previously unseen records. Our findings revealed that the proposed strategy of dataset augmentation, combined with techniques such as dynamic learning rate adjustment and early stopping, not only optimizes computational efficiency but also achieves superior classification performance. 

The future research direction can focus on:
\begin{itemize}
    \item Advanced ensemble techniques: Use methods like stacking or boosting to improve classification accuracy.
    \item Multimodal data integration: Include more academic domains and incorporate metadata or citation networks for better context-aware classification.
    \item Enhance context understanding: Enhance models to capture relationships and new terms in specific domains.
    \item Cross-domain testing: Test models on new but related domains to improve adaptability.
\end{itemize}

\section{Limitations}
This study, while contributing valuable insights into domain-specific text classification, faced several limitations:
\begin{itemize}
    \item Domain-specific models like SciBERT, and BioBERT achieved remarkable success in certain fields but underperformed in domains like civil engineering and psychology.
    \item Despite dataset augmentation, some domains with sparse data, like medical sciences, posed challenges in achieving balanced performance.
    \item Fine-tuning large language models required significant computational power, limiting the approach's scalability and practical application.
\end{itemize}

\section*{Acknowledgements}
The authors express their gratitude to the members of the Applied Machine Learning Research Group at Obuda University's John von Neumann Faculty of Informatics for their valuable comments and suggestions. They also wish to acknowledge the support provided by the Doctoral School of Applied Informatics and Applied Mathematics at Obuda University.
\bibliographystyle{ieeetr}
\bibliography{ref}

\end{document}